\pdfoutput=1

\documentclass[11pt]{article}

\usepackage{EMNLP2023}

\usepackage{times}
\usepackage{latexsym}

\usepackage[T1]{fontenc}

\usepackage[utf8]{inputenc}

\usepackage{microtype}

\usepackage{inconsolata}

\usepackage{amsmath}
\usepackage{multirow}
\usepackage{graphicx}
\usepackage{caption}
\usepackage{subcaption}
\usepackage{amssymb} 
\usepackage{xcolor} 

\title{Integrating Language Models into Direct Speech Translation: \\
An Inference-Time Solution to Control Gender Inflection}

\author{\textbf{Dennis Fucci},\textsuperscript{1,2} \textbf{Marco Gaido},\textsuperscript{2} \textbf{Sara Papi},\textsuperscript{1,2} \\ \textbf{Mauro Cettolo},\textsuperscript{2} \textbf{Matteo Negri},\textsuperscript{2} \textbf{Luisa Bentivogli}\textsuperscript{2}\\
  \textsuperscript{1}University of Trento\\
  \textsuperscript{2}Fondazione Bruno Kessler\\
  {\tt \{dfucci,mgaido,spapi,cettolo,negri,bentivo\}@fbk.eu} \\}
  
\begin{document}
\maketitle
\begin{abstract}

When translating words referring to the speaker, speech translation (ST) systems should not resort to default masculine generics nor rely on potentially misleading vocal traits.  Rather, they should assign gender according to the speakers' preference. 
The existing solutions to do so, though effective, 
are hardly feasible in practice
as they involve dedicated model re-training 
on gender-labeled ST data.
To overcome these limitations,
we propose the first inference-time solution to control speaker-related gender inflections in ST. Our approach partially replaces the (biased) \textit{internal} language model (LM) implicitly learned by 
the ST decoder
with gender-specific \textit{external} LMs. 
Experiments on en$\rightarrow$es/fr/it show that our 
solution outperforms the base models and the best training-time mitigation strategy by up to 31.0 and 1.6 points in gender accuracy, respectively, for feminine forms. 
The gains are even larger (up to 32.0 and 3.4) in the challenging condition where speakers' vocal traits conflict with their 
gender.\footnote{\label{foot:gender_disclaimer}Note that, throughout the paper, when using the terms \textit{female, male,} and \textit{gender} we do not refer to speakers' gender identity but exclusively to their preferred linguistic expression of gender (see \S\ref{sec:impact} for an in-depth discussion of this issue).}

\end{abstract}

\section{Introduction}
\label{sec:intro}

The problem of gender bias in automatic translation particularly emerges when translating from genderless or notional gender languages (e.g., English) -- which feature limited gender-specific marking -- into grammatical gender languages (e.g., Spanish) -- which exhibit a rich lexical and morpho-syntactic system of gender \citep{savoldi-etal-2021-gender}.
In this scenario, 
when gender-neutral words are translated into gender-marked words
(e.g. 
en: \textit{the nurse} -- 
es:
\textit{\underline{el}/\underline{la} enfermer\underline{o}/\underline{a}}), both machine translation (MT) and speech translation (ST) systems are often biased towards masculine or stereotypical predictions 
\citep{cho-etal-2019-measuring,Prates2018AssessingGB,bentivogli-etal-2020-gender,CostaJussa-GenderArchitecture},
especially in absence of explicit cues 
(en: \textit{the nurse and \underline{his} dog}).
A common instance is represented by words that refer to the first-person subject (henceforth referred to as speaker-dependent words, such as \textit{I'm \underline{a young nurse}}).
In these cases, direct ST systems \cite{directST} have been shown to rely on vocal 
traits to determine gender inflections \cite{bentivogli-etal-2020-gender}.
This, however, 
does not eliminate the bias toward masculine forms and
is not
inclusive for 
those individuals whose vocal properties do not align with their
gender,
such as
people with vocal impairments, children, and transgenders
\citep{matar,MENEZES2022}.
Therefore, whenever the 
speaker's gender\textsuperscript{\ref{foot:gender_disclaimer}} 
is known (e.g. in talks or lectures), 
such information should be 
exploited
to control gender 
translation and 
avoid relying on potentially misleading
physical cues.

So far, this topic
has been 
investigated only by \citet{gaido-etal-2020-breeding}.
Their best solution consists in creating two gender-specific \emph{specialized} models by fine-tuning a generic direct ST system on sentences 
uttered by female/male speakers.
Though effective, this 
method
has inherent limitations. 
First, 
it
requires parallel audio-text data labeled with speakers' gender, which are scarcely available and costly to collect.
Second, 
the fine-tunings
are computationally demanding as they involve processing 
audio data,
which are much longer ($\sim$8$\times$) than
their
textual equivalents \cite{salesky-etal-2019-exploring}.

To overcome these limitations, we propose the first inference-time solution in direct ST to control gender translation for speaker-dependent words when the speaker's gender is known.\footnote{
Code and models available at \url{https://github.com/hlt-mt/FBK-fairseq} under Apache License 2.0.}
Our approach 
guides
gender translation by partially substituting the biased \textit{internal language model} 
implicitly learned by the ST decoder of a base model with a gender-specific \textit{external language model} 
learned on monolingual textual data.
Through experiments on three language pairs (en$\rightarrow$es/fr/it), 
we demonstrate that, in terms of gender accuracy, our solution
outperforms the base system by up to 31.0 points (for feminine forms) and is on par with the best training-time approach (with up to 1.6 of gain for feminine forms). Its effectiveness is also confirmed when speakers’ vocal traits conflict with their gender, with 
gains 
up to 32.0 and 3.4
over the base system and the best training-time solution.

\section{ILM/ELM for Gender Translation}
\label{sec:method}

The autoregressive decoder of an encoder-decoder architecture is trained to predict 
the next target token given the previous ones and the encoder output.
Thereby,
it implicitly learns to model the target language from the training data, thus developing an \textit{internal language model} (ILM)  \citep{McDermott2019ADR,VarianiHAT}.
We assume that, in a direct ST model trained on unbalanced data where female speakers (and consequently feminine speaker-dependent words) are under-represented \cite{tatman-2017-gender}, the ILM is biased toward masculine forms. Therefore, we propose to guide the generation of the ST model with respect to speaker-dependent words  by 
substituting the biased ILM with a gender-specific \textit{external language model} (ELM).
To this aim, we train two ELMs on
monolingual text corpora
(easy
to collect,
unlike labelled audio data)
containing either
feminine or masculine
speaker-dependent words (see \S\ref{sec:data}). 
At inference time,
when we have prior knowledge of the speaker's gender from the metadata,
we \textit{i)} integrate the ELM specialized in either masculine or feminine forms (depending on the speaker's gender)
into the ST model,
and \textit{ii)} (partially) remove the ILM contribution.

The integration of end-to-end models with ELMs is a widespread solution to leverage text data in speech recognition 
\citep{Bahdanau2016, Chorowski2016TowardsBD, Kannan2018, irie19b_interspeech}.
Successful applications span from recognizing rare words \citep{Sainath2021AnES, Huang2022SentenceSelectLL}  to coping with out-of-vocabulary terms \citep{Hori2017}, domain adaptation \citep{sriram,Shan} and under-resourced conditions \citep{McDermott2019ADR}.
However, to the best of our knowledge, ELM integration has not been explored in the field of direct ST, nor in the context of gender translation, as we do here.
Among the various methods proposed 
for the ELM integration \citep{gulchere2015, gulchere2017, sriram, stahlberg, Shan, McDermott2019ADR},
we avoid those that require training-time interventions, and we resort to \textit{shallow fusion} \citep{gulchere2015, gulchere2017},  
an effective technique
\citep{Kannan2018, Inaguma_2019_transfer}
that
consists in the log-linear combination
of the posterior of the base model ($p_{M_B}$) and the prior of the ELM ($p_{ELM}$).

As regards the ILM removal, 
which previous studies already 
shown
to amplify the performance gains yielded by ELM integration \citep{Meng-ICASSP,Meng-SLT,Meng-Interspeech,andresferrer21_interspeech,liu22j_interspeech,10095249},
the most critical aspect is 
its estimation.
In fact, since the ILM is implicitly modeled in the decoder, disentangling its contribution from the rest of the network is a challenging 
task
\cite{VarianiHAT, Meng-ICASSP, Meng-SLT, Zeineldeen}.
Among the estimation methods demonstrated by \citet{Zeineldeen} to yield the best results, we select the \textit{global encoder average}, as it does not require training-time interventions.
This method computes the ILM prior ($p_{ILM}$) as:
\begin{equation}
\nonumber
\begin{gathered}\label{eqn:ilm}
p_{ILM}(y) = p_{M_{B_{decoder}}}(y|c)
\end{gathered}
\end{equation}
namely, by feeding the ST decoder with the average $c$ of the encoder outputs $h_{n,t}$ over all the $T_n$ timesteps of the $N$ training samples, where $c$ is:
\begin{equation}
\nonumber
\begin{gathered}\label{eqn:c}
c = \frac{1}{\sum_{n=1}^{N}T_{n}} \sum_{n=1}^{N} \sum_{t=1}^{T_{n}}h_{n,t}
\end{gathered}
\end{equation}

Therefore, given an audio input $x$, the output $\widehat{\mathrm{y}}$ of our solution
is the translation $y$ that maximizes the log-linear combination of $p_{M_B}$, $p_{ELM}$ and $p_{ILM}$:
\begin{gather*}
    \widehat{\mathrm{y}} =\underset{y}{\mathrm{argmax}}\{\log p_{M_B}(y|x) - \beta_{ILM} \log p_{ILM}(y) \\
    + \beta_{ELM}\log p_{ELM}(y)\}
\end{gather*}
%
%
where $\beta_{ILM}$ and $\beta_{ELM}$ are positive scalar weights
calibrating 
ELM
integration and ILM removal.

\begin{table*}[tb]
\centering
\small
\setlength{\tabcolsep}{4.0pt}
\begin{tabular}{l|cc|cc||cc|cc||cc|cc}
\cline{1-13} 
\multirow{3}{*}{} & \multicolumn{4}{c||}{\textbf{es}} & \multicolumn{4}{c||}{\textbf{fr}} & \multicolumn{4}{c}{\textbf{it}} \\
\cline{2-13} 
& \multicolumn{2}{c|}{\textbf{train}} & \multicolumn{2}{c||}{\textbf{dev}} & \multicolumn{2}{c|}{\textbf{train}} & \multicolumn{2}{c||}{\textbf{dev}} & \multicolumn{2}{c|}{\textbf{train}} & \multicolumn{2}{c}{\textbf{dev}} \\
\cline{2-13} 
 & M & F & M & F & 
 M & F & M & F & 
 M & F & M & F \\
\cline{1-13}
\multicolumn{1}{l|}{Sent.} & 
196.8K & 111.9K & 1.6K & 1.2K &
566.9K & 232.4K & 8.5K & 3.3K & 
370.7K & 171.9K & 5.3K & 3.0K \\
\multicolumn{1}{l|}{Words} & 
4.1M & 2.4M & 37.5K & 26.7K & 
13.7M & 5.5M & 232.2K & 87.1K & 
8.9M & 4.2M & 132.3K & 75.4K  \\
\cline{1-13}
\end{tabular}
\caption{Statistics for the monolingual text corpora collected.}
\label{tab:n-data}
\end{table*}

The three components 
($p_{M_B}$, $p_{ELM}$, and $p_{ILM}$) 
convey different 
information:
\textit{i)} $p_{M_B}$ embeds both the acoustic and the linguistic information learned from the ST data; 
\textit{ii)} $p_{ILM}$ represents the estimated linguistic knowledge learned by $M_{B}$; \textit{iii)} $p_{ELM}$ embeds linguistic information 
(in our case gender-specific forms) 
learned from external textual resources. 
Therefore,
$\beta_{ILM}$ and $\beta_{ELM}$ must be set to values that effectively integrate the
internal and external linguistic knowledge, so that the gender bias affecting the ST decoder is mitigated 
by the ELM.
At the same time, the linguistic contribution supplied by the ELM must not override the acoustic modelling capabilities of $p_{M_B}$, so as to avoid 
translation quality drops. 
Accordingly, we estimate $\beta_{ILM}$ and $\beta_{ELM}$ by optimizing the harmonic mean of the two metrics (gender accuracy and BLEU -- see $\S$\ref{sec:data}) used to measure gender bias and overall translation quality, so as to equally weigh our two objectives.
In Appendix \ref{app:beta}, we discuss the 
computation of  $\beta_{ILM}$ and $\beta_{ELM}$ values,
also showing that their precise estimation is not critical since final results are rather robust to small weight variations.

\section{Data and Metrics}
\label{sec:data}

Our 
en$\rightarrow$es/fr/it 
ST systems are trained on the TED-based
MuST-C corpus \cite{CATTONI2021101155}.
This resource includes a manual annotation of the speakers’ gender \cite{gaido-etal-2020-breeding}, which is used to determine the gender translation of speaker-dependent words.
To train the ELMs, 
we collected 
GenderCrawl,\footnote{
Available at \url{https://mt.fbk.eu/gendercrawl/} under the Creative Commons Attribution 4.0 International license (CC BY 4.0).}
a set of 
monolingual corpora
for each target language and
gender.
Each corpus is made of sentences 
with speaker-dependent words that clarify the speaker's gender
(e.g., 
es: \textit{Soy \underline{nueva}} <F> \textit{en esta zona} [en: \textit{I am new to this area}], es: \textit{Debía ser fiel a mi \underline{mismo}} <M> [en: \textit{I had to be true to myself}]).
These sentences were automatically selected from ParaCrawl \cite{banon-etal-2020-paracrawl}
through regular expressions representative of morpho-syntactic patterns 
matching 
references to the first-person singular.
Additionally, we have also collected a validation set by applying the same regular expressions to the MuST-C training sets.
The statistics of all these datasets are presented in Table \ref{tab:n-data}.

We evaluate our systems on the
TED-derived and gender-sensitive MuST-SHE benchmark \cite{bentivogli-etal-2020-gender}. 
In particular, we focus on its ``Category 1'', which contains 
from 560 to 607 sentences (depending on the target language)
with speaker-dependent words annotated in the reference.
%
%
To assess gender translation, we use
the official MuST-SHE evaluation script\footnote{\url{https://mt.fbk.eu/must-she/}.}, which
produces two measures:  
\textit{i)}~\textit{term coverage}, i.e. the 
percentage
of annotated words that are generated by the system (disregarding their gender marking), and on which gender translation is hence automatically measurable,
and \textit{ii)} \textit{gender accuracy}, i.e. the 
percentage
of words generated in the correct gender among the measurable ones.
Lastly,
overall translation quality is calculated 
with SacreBLEU~\cite{post-2018-call}.\footnote{case:mixed$\vert$eff:no$\vert$tok:13a$\vert$smooth:exp$\vert$version:2.0.0}

\begin{table*}[tb]
\centering
\small
\setlength{\tabcolsep}{2.2pt}
\begin{tabular}{l||c|cc|cc||c|cc|cc||c|cc|cc} 
\hline
\multirow{3}{*}{\textbf{Models}} & \multicolumn{5}{c||}{\textbf{en-es}} & \multicolumn{5}{c||}{\textbf{en-fr}} & \multicolumn{5}{c}{\textbf{en-it}} \\ 
\cline{2-16}
 & \multirow{2}{*}{\textbf{BLEU}} & \multicolumn{2}{c|}{\textbf{Coverage}} & \multicolumn{2}{c||}{\textbf{Gender Acc.}} &
\multirow{2}{*}{\textbf{BLEU}} & \multicolumn{2}{c|}{\textbf{Coverage}} & \multicolumn{2}{c||}{\textbf{Gender Acc.}} &
 \multirow{2}{*}{\textbf{BLEU}} & \multicolumn{2}{c|}{\textbf{Coverage}} & \multicolumn{2}{c}{\textbf{Gender Acc.}} \\ 
 &  & M & F & M & F & 
 & M & F & M & F & 
 & M & F & M & F \\ 
\hline
M\textsubscript{B} & 
34.8 & 
65.1 & 67.9 & 
71.6 & 45.7 &
29.8 & 
51.5 & 55.9 & 
72.5 & 52.0 & 
26.8 & 
51.6 & 50.6 & 
77.3 & 49.5 \\
M\textsubscript{SP} & 
\textbf{35.2} & 
64.8 & 66.8 & 
\textbf{85.6} & \textbf{76.8} & 
\textbf{29.9} & 
52.9 & 55.2 & 
\textbf{92.4} & 78.5 & 
\textbf{27.3} & 
\textbf{52.8} & 49.4 & 
\textbf{92.5} & 73.3 \\
M\textsubscript{B+ELM} & 
33.7\textsuperscript{ab} & 
\textbf{67.5}\textsuperscript{AB} & 68.1 &
77.9\textsuperscript{Ab} & 69.2\textsuperscript{Ab} & 
29.3 & 
\textbf{54.9}\textsuperscript{A} & \textbf{57.1} & 
81.9\textsuperscript{Ab} & 75.8\textsuperscript{A} & 
27.2 & 
51.8 & \textbf{54.4}\textsuperscript{AB} & 
81.2\textsuperscript{Ab} & 72.8\textsuperscript{A} \\
M\textsubscript{B-ILM+ELM} & 
34.4\textsuperscript{b} & 
65.8 & \textbf{71.2}\textsuperscript{AB} & 
82.3\textsuperscript{A} & 76.7\textsuperscript{A} & 
29.8 & 
54.4\textsuperscript{A} & 56.1 & 
84.5\textsuperscript{Ab} & \textbf{79.2}\textsuperscript{A} & 
27.2 & 
52.3 & 54.1\textsuperscript{AB} & 
84.9\textsuperscript{Ab} & \textbf{74.9}\textsuperscript{A} \\
\hline
\end{tabular}
\caption{BLEU ($\uparrow$), (term) coverage ($\uparrow$), and M/F gender accuracy (Gender Acc., $\uparrow$) scores.
\textsuperscript{A/a} and \textsuperscript{B/b} indicate that the improvement (uppercase) or the degradation (lowercase) of our technique over the baseline (M\textsubscript{B}) and the fine-tuning approach (M\textsubscript{SP}), respectively, 
is statistically significant (bootstrap resampling with 95\% CI, \citealt{koehn-2004-statistical}).}
\label{tab:results-all}
\end{table*}

\section{Results}
\label{sec:results}

For each language pair, we evaluate our approach by training:
\textit{i)} an ST baseline model (\textbf{M\textsubscript{B}}) that is not aware of the 
speaker’s
gender;
\textit{ii)}~the specialized models (\textbf{M\textsubscript{SP}}) presented in \cite{gaido-etal-2020-breeding}, re-implemented as upper bound to compare our inference-time solution with the best training-time approach;
\textit{iii)} the 
combination
of M\textsubscript{B} with the gender-specific ELMs and the ILM removal 
(\textbf{M\textsubscript{B-ILM+ELM}});
\textit{iv)} a variant of 
the approach,
where the ILM is not removed (\textbf{M\textsubscript{B+ELM}}),
serving as an ablation study to disentangle the ILM and ELM contributions.
Detailed experimental settings and model description are provided in Appendix \ref{app:models}.

\subsection{Main Results}
\label{sec:main_results}

Table \ref{tab:results-all} presents 
BLEU, term coverage, and gender accuracy scores for all language pairs,
divided into 
feminine/masculine (F/M)
forms.

\noindent\textbf{Gender Accuracy.}
The results indicate that our approach, both with and without the ILM removal, significantly outperforms M\textsubscript{B} on all language pairs. 
Specifically, M\textsubscript{B-ILM+ELM} is always better than M\textsubscript{B+ELM}, 
demonstrating
that the 
ILM removal in combination with ELM integration improves debiasing.
The accuracy gains of M\textsubscript{B-ILM+ELM} over M\textsubscript{B} are particularly high on feminine forms, ranging from 25.4 to 31.0. 
In addition, the accuracy
of M\textsubscript{B-ILM+ELM} is comparable to that of the training-time approach M\textsubscript{SP}. 
While 
M\textsubscript{SP} is significantly superior only for M in en-it and en-fr, 
M\textsubscript{B-ILM+ELM} is the best on average for F, the most misgendered category.

\noindent\textbf{Translation Quality.} 
Looking at
BLEU scores, we notice that, with the only exception of en-it, the simple integration of the ELM 
(M\textsubscript{B+ELM}) degrades the quality with respect to both M\textsubscript{B} and M\textsubscript{SP},\footnote{In \citet{gaido-etal-2020-breeding}, the \textit{specialized} 
systems achieve higher results
as their base models are built using large ST, ASR, and MT corpora, while we train only on MuST-C.}
especially in en-es where the drops are statistically significant. 
The ILM 
removal mostly solves the problem, as M\textsubscript{B-ILM+ELM} achieves scores that are comparable to 
M\textsubscript{SP} 
on en-fr and en-it, 
and partly closes the gap on en-es, where the drop with respect to M\textsubscript{B} (-0.4) is not statistically significant. 
Interestingly, looking at term
coverage, 
both M\textsubscript{B-ILM+ELM} and M\textsubscript{B+ELM} consistently outperform M\textsubscript{B} and M\textsubscript{SP}, 
with the only exception of masculine words in en-it. 
In particular, the gains are high for feminine words, where M\textsubscript{B-ILM+ELM} significantly outperforms both M\textsubscript{B} and M\textsubscript{SP}. 
This shows 
that the integration of 
textual data can 
increase the ability to model feminine vocabulary,
less represented in training data.

In conclusion, our inference-time solution effectively
improves gender translation
in direct ST, 
especially for feminine forms (see Appendix \ref{app:examples} for output examples). Moreover, it achieves comparable results with the best training-time approach, while overcoming its limitations. 
Such improvements do not come at the detriment of the overall translation quality (as shown by BLEU scores) nor of the accuracy in 
assigning gender to
words 
that pertain to human referents other than the speaker 
(as shown in Appendix \ref{app:cat2}).

\subsection{Robustness to Vocal Traits}
\label{subsec:robustness}
We also evaluate
the inclusivity of our solution for speakers whose vocal traits are stereotypically 
associated with a gender opposite to their own.
As 
MuST-SHE solely 
contains
utterances
from speakers whose gender aligns with 
their vocal properties,
we 
simulate
this condition 
using the provided
``wrong references'',
in which the
speaker-dependent words 
are
swapped to the opposite gender.
We 
treat
them as correct references, so as to have female voices with masculine targets and vice versa, 
and we require the systems to produce the output with the gender of the target.
Table \ref{tab:counterfactual} shows BLEU, term coverage, and gender accuracy 
for
M\textsubscript{B}, M\textsubscript{SP}, and our best-performing model M\textsubscript{B-ILM+ELM},
averaged over the  
three language pairs.

\noindent\textbf{Gender Accuracy.} 
Regarding gender realization, M\textsubscript{B-ILM+ELM} performs 
noticeably
better than M\textsubscript{B}, as we observe
a substantial improvement of 19.7 points in 
producing masculine forms (Voice F--Gdr M) 
and 32.0 in 
producing feminine forms (Voice M--Gdr F). 
This suggests
that our approach is capable of
partially overriding the vocal information, 
on which the
base model
unduly
relies to translate the 
speaker-dependent words.
In comparison with 
M\textsubscript{SP}, 
our approach is inferior in 
Voice F--Gdr M,
while it is superior in generating 
the less-represented
feminine translation (Voice M--Gdr F), confirming the trends observed in the previous scenario (see \S\ref{sec:main_results}).

\noindent\textbf{Translation Quality.} 
In terms of BLEU, 
our approach (M\textsubscript{B-ILM+ELM}) is on par with the training-time strategy (M\textsubscript{SP}),
but they both suffer a 
$\sim$2.5 BLEU drop with respect to the base system (M\textsubscript{B}).
The reason for this drop may lay on the fact that gender-specific models learned patterns
that differentiate male and female language \cite{Mulac2001,boulis-ostendorf-2005-quantitative}, 
which are disregarded when only swapping the gendered words in the references.
However, M\textsubscript{B-ILM+ELM} 
outperforms M\textsubscript{B} and M\textsubscript{SP} in terms of coverage, 
with a marginal gain (0.5-0.6) for 
male
speakers (Voice M--Gdr F) and a 
larger
gain (2.3-3.7) for 
female speakers (Voice F--Gdr M), 
confirming that our approach increases 
the coverage of the vocabulary used by females (even when expressed in the masculine form).

All in all, the experiments in this challenging testing condition prove that our solution effectively overrides the reliance of base ST systems on speakers' vocal traits. Also, they confirm its superiority in translating the less-represented feminine forms.

\begin{table}[!t]
\small
\centering
\setlength{\tabcolsep}{2.8pt}
\begin{tabular}{l||c|cc|cc} 
\hline
\multirow{3}{*}{\textbf{Models}} & \multicolumn{5}{c}{\textbf{Average}} \\ 
\cline{2-6}
 & \multirow{3}{*}{\textbf{BLEU}} & \multicolumn{2}{c|}{\textbf{Coverage}} & \multicolumn{2}{c}{\textbf{Gender Acc.}} \\ 
  &   & Voice F & Voice M & Voice F & Voice M \\ 
  &   & Gdr M & Gdr F & Gdr M & Gdr F \\ 
\hline
M\textsubscript{B} &
\textbf{30.5} & 
58.0 & 56.2 & 
50.9 & 26.1 \\
M\textsubscript{SP} & 
28.2 & 
56.6 & 56.1 & 
\textbf{83.0} & 54.7 \\
M\textsubscript{B-ILM+ELM} & 
28.0 & 
\textbf{60.3} & \textbf{56.7} & 
70.6 & \textbf{58.1} \\
\hline
\end{tabular}
\caption{BLEU, term coverage, and gender accuracy for the conflicting scenario averaged over en$\rightarrow$es/fr/it.} 
\label{tab:counterfactual}
\end{table}

\section{Conclusions}

We proposed the first inference-time solution to control gender translation of speaker-dependent words in direct ST.
Our approach partially replaces the biased ILM of the ST decoder with a gender-specific ELM. As such, it can be applied to existing models without 
the need for
labeled ST data or computationally expensive re-trainings,
overcoming the limitations of existing training-time methods.
Experiments on three language pairs 
proved the effectiveness of our technique in controlling gender inflections of 
words referring to the first-person subject,
regardless of whether the speakers' vocal traits are aligned with their gender or not.
In addition to significantly increasing the gender accuracy of base ST models, it achieves substantial parity with the best training-time method while consistently increasing the correct generation of feminine forms.

\section{Acknowledgements}
This work is part of the project “Bias Mitigation and Gender Neutralization Techniques for Automatic Translation”, which is financially supported by an Amazon Research Award AWS AI grant. 
Moreover, we acknowledge the support of the PNRR project FAIR - Future AI Research (PE00000013), under the NRRP MUR program funded by the NextGenerationEU.

\section{Limitations}
\label{sec:limitations}

In our experiments, 
we exclusively evaluated our approach on English to Romance language translations. 
Conducting experiments on different language pairs would be valuable. 
However, it is important to note that such endeavors would demand substantial efforts in annotating data, as benchmarks akin to MuST-SHE are currently unavailable for other target languages.

Our inference-time solution, as described in the paper, 
significantly reduces 
the computational costs of current
approaches
by eliminating the need for ST retraining.
However, there is 
an increase in inference costs, due to the additional forward passes on the ELM and ILM 
(which is the same as the ST decoder, but fed with a different encoder output).
In particular, 
since our implementation has not been optimized and performs the operations sequentially,
our solution reduces the inference speed (computed as the number of generated tokens per second) by 
$\sim$40\% (from 165 to 100).\footnote{Statistics computed on a p3.2xlarge instance on AWS (featuring one NVIDIA V100 GPU).} 
Such slowdown can be reduced by: \textit{i)} parallelizing the forward passes of the ST model, ELM, and ILM; \textit{ii)} 
caching computed states in the ILM to avoid recomputation at each generation step.
Optimizing our implementation,
although necessary for production usage,
is outside the scope of our work.

Lastly,
our ELM implementation uses
the
same
BPE \citep{sennrich-etal-2016-neural} vocabulary of 
the ST models, 
trained
on the
textual target of
MuST-C.
Due to the under-representation of feminine forms
in this corpus,
statistical segmentation methods like BPE split the less frequent feminine forms into less compact sequences of tokens (for example, in our experiments, we observed the split \textit{maes\_tra} vs \textit{maestro} for Spanish).
This tokenization process can penalize generalization on morphology and, consequently, gender translation when compared to character-level representations \citep{belinkov-coli}.
As such, an interesting future direction
is represented by training the ELMs with a character-based vocabulary, which
has the potential to
enhance gender accuracy
and 
further increase
the significant gains
already
achieved.

\section{Ethics Statement}
\label{sec:impact}

In this paper we 
presented a new methodology to 
improve
ST systems in their ability to correctly generate masculine and feminine forms for
first-person-singular referents. 
Hereafter, we contextualize the impact of our research and discuss the ethical principles at the basis of our work.

We define gender bias in MT/ST
as the tendency of systems to systematically favor masculine forms 
to the detriment of
the feminine ones when related to human entities \citep{crawford2017trouble}.
This bias not only hampers the 
performance
of the system
by producing erroneous translations of gender-marked words, 
but also has significant societal implications. 
For example, incorrect gender translations 
can impact self-perception, 
as linguistic expressions of gender play a crucial role in negotiating and communicating personal representation 
\cite{Stahlberg-gender, corbett, gygax}.
According to \citet{blodgett-etal-2021-stereotyping} and \citet{savoldi-etal-2021-gender}, 
gender bias in translation technologies leads to both \textit{representational harms}, 
such as under-representation of women and diminished visibility of their linguistic repertoire, 
and \textit{allocational harms}, characterized by unequal quality of service due to performance disparities between male and female users.

In light of the above, we believe that our solution
positively impacts single individuals
and
society at large, by improving
not only
the experience of using such technologies
but also
feminine visibility.
Furthermore,
by relying on explicit gender information, 
our mitigation solution goes beyond 
a mere and potentially misleading 
exploitation of
the speech signal.
Indeed, 
using 
speaker's vocal properties
would foster the stereotypical expectations about how masculine or feminine voices should sound, 
which
is not inclusive for certain users, 
such as transgender individuals or people with laryngeal diseases \cite{matar, pereira, VillasBoas, MENEZES2022}.

As regards possible concerns about the gender information considered in our experiments,
we relied on the annotations of the two datasets used, MuST-C/MuST-Speakers and MuST-SHE. Both these resources have been manually annotated with speakers' gender information based on the personal pronouns found in their public TED profile \cite{gaido-etal-2020-breeding, bentivogli-etal-2020-gender}. We follow the statement of the curators of these resources, thus bearing in mind that the gender tag accounts only for the linguistic gender by which the speakers accept to be referred to in English
and to which they would like the translation to conform.
We acknowledge that this information does not necessarily correspond to the speakers’ self-determined gender identity \cite{cao-daume-iii-2020-toward}.
We are also aware that we cannot consider their preference as static in time \cite{hovy}.

Last but not least, in this work
we only consider binary linguistic forms
as they are the only ones represented in the currently available ST data.
In fact, to the best of our knowledge, ST corpora also representing non-binary speakers are not yet available. However, we encourage a vision of gender going beyond binarism
and we believe that extending the application of our method to non-binary forms (e.g. by integrating a third, \textit{non-binary} ELM)
can be an interesting extension of this work.


\bibliographystyle{acl_natbib}
\bibliography{anthology, custom}

\begin{thebibliography}{61}
\expandafter\ifx\csname natexlab\endcsname\relax\def\natexlab#1{#1}\fi

\bibitem[{Andr{\'{e}}s{-}Ferrer et~al.(2021)Andr{\'{e}}s{-}Ferrer, Albesano,
  Zhan, and Vozila}]{andresferrer21_interspeech}
Jes{\'{u}}s Andr{\'{e}}s{-}Ferrer, Dario Albesano, Puming Zhan, and Paul
  Vozila. 2021.
\newblock \href {https://doi.org/10.21437/Interspeech.2021-443} {Contextual
  density ratio for language model biasing of sequence to sequence {ASR}
  systems}.
\newblock In \emph{Interspeech 2021, 22nd Annual Conference of the
  International Speech Communication Association, Brno, Czechia, 30 August - 3
  September 2021}, pages 2007--2011. International Speech Communication
  Association.

\bibitem[{Bahdanau et~al.(2016)Bahdanau, Chorowski, Serdyuk, Brakel, and
  Bengio}]{Bahdanau2016}
Dzmitry Bahdanau, Jan Chorowski, Dmitriy Serdyuk, Philémon Brakel, and Yoshua
  Bengio. 2016.
\newblock \href {https://doi.org/10.1109/ICASSP.2016.7472618} {End-to-end
  attention-based large vocabulary speech recognition}.
\newblock In \emph{2016 IEEE International Conference on Acoustics, Speech and
  Signal Processing (ICASSP)}, pages 4945--4949. Institute of Electrical and
  Electronics Engineers.

\bibitem[{Ba{\~n}{\'o}n et~al.(2020)Ba{\~n}{\'o}n, Chen, Haddow, Heafield,
  Hoang, Espl{\`a}-Gomis, Forcada, Kamran, Kirefu, Koehn, Ortiz~Rojas,
  Pla~Sempere, Ram{\'\i}rez-S{\'a}nchez, Sarr{\'\i}as, Strelec, Thompson,
  Waites, Wiggins, and Zaragoza}]{banon-etal-2020-paracrawl}
Marta Ba{\~n}{\'o}n, Pinzhen Chen, Barry Haddow, Kenneth Heafield, Hieu Hoang,
  Miquel Espl{\`a}-Gomis, Mikel~L. Forcada, Amir Kamran, Faheem Kirefu, Philipp
  Koehn, Sergio Ortiz~Rojas, Leopoldo Pla~Sempere, Gema
  Ram{\'\i}rez-S{\'a}nchez, Elsa Sarr{\'\i}as, Marek Strelec, Brian Thompson,
  William Waites, Dion Wiggins, and Jaume Zaragoza. 2020.
\newblock \href {https://doi.org/10.18653/v1/2020.acl-main.417} {{P}ara{C}rawl:
  Web-scale acquisition of parallel corpora}.
\newblock In \emph{Proceedings of the 58th Annual Meeting of the Association
  for Computational Linguistics}, pages 4555--4567, Online. Association for
  Computational Linguistics.

\bibitem[{Belinkov et~al.(2020)Belinkov, Durrani, Dalvi, Sajjad, and
  Glass}]{belinkov-coli}
Yonatan Belinkov, Nadir Durrani, Fahim Dalvi, Hassan Sajjad, and James Glass.
  2020.
\newblock \href {https://doi.org/10.1162/coli_a_00367} {{On the Linguistic
  Representational Power of Neural Machine Translation Models}}.
\newblock \emph{Computational Linguistics}, 46(1):1--52.

\bibitem[{Bentivogli et~al.(2020)Bentivogli, Savoldi, Negri, Di~Gangi, Cattoni,
  and Turchi}]{bentivogli-etal-2020-gender}
Luisa Bentivogli, Beatrice Savoldi, Matteo Negri, Mattia~A. Di~Gangi, Roldano
  Cattoni, and Marco Turchi. 2020.
\newblock \href {https://doi.org/10.18653/v1/2020.acl-main.619} {Gender in
  danger? evaluating speech translation technology on the {M}u{ST}-{SHE}
  corpus}.
\newblock In \emph{Proceedings of the 58th Annual Meeting of the Association
  for Computational Linguistics}, pages 6923--6933, Online. Association for
  Computational Linguistics.

\bibitem[{Blodgett et~al.(2021)Blodgett, Lopez, Olteanu, Sim, and
  Wallach}]{blodgett-etal-2021-stereotyping}
Su~Lin Blodgett, Gilsinia Lopez, Alexandra Olteanu, Robert Sim, and Hanna
  Wallach. 2021.
\newblock \href {https://doi.org/10.18653/v1/2021.acl-long.81} {Stereotyping
  {N}orwegian salmon: An inventory of pitfalls in fairness benchmark datasets}.
\newblock In \emph{Proceedings of the 59th Annual Meeting of the Association
  for Computational Linguistics and the 11th International Joint Conference on
  Natural Language Processing (Volume 1: Long Papers)}, pages 1004--1015,
  Online. Association for Computational Linguistics.

\bibitem[{Boulis and Ostendorf(2005)}]{boulis-ostendorf-2005-quantitative}
Constantinos Boulis and Mari Ostendorf. 2005.
\newblock \href {https://doi.org/10.3115/1219840.1219894} {A quantitative
  analysis of lexical differences between genders in telephone conversations}.
\newblock In \emph{Proceedings of the 43rd Annual Meeting of the Association
  for Computational Linguistics ({ACL}{'}05)}, pages 435--442, Ann Arbor,
  Michigan. Association for Computational Linguistics.

\bibitem[{Bérard et~al.(2018)Bérard, Besacier, Kocabiyikoglu, and
  Pietquin}]{directST}
Alexandre Bérard, Laurent Besacier, Ali~Can Kocabiyikoglu, and Olivier
  Pietquin. 2018.
\newblock \href {https://doi.org/10.1109/ICASSP.2018.8461690} {End-to-end
  automatic speech translation of audiobooks}.
\newblock In \emph{2018 IEEE International Conference on Acoustics, Speech and
  Signal Processing (ICASSP)}, pages 6224--6228. Institute of Electrical and
  Electronics Engineers.

\bibitem[{Cao and Daum{\'e}~III(2020)}]{cao-daume-iii-2020-toward}
Yang~Trista Cao and Hal Daum{\'e}~III. 2020.
\newblock \href {https://doi.org/10.18653/v1/2020.acl-main.418} {Toward
  gender-inclusive coreference resolution}.
\newblock In \emph{Proceedings of the 58th Annual Meeting of the Association
  for Computational Linguistics}, pages 4568--4595, Online. Association for
  Computational Linguistics.

\bibitem[{Cattoni et~al.(2021)Cattoni, {Di Gangi}, Bentivogli, Negri, and
  Turchi}]{CATTONI2021101155}
Roldano Cattoni, Mattia~Antonino {Di Gangi}, Luisa Bentivogli, Matteo Negri,
  and Marco Turchi. 2021.
\newblock \href {https://doi.org/https://doi.org/10.1016/j.csl.2020.101155}
  {Must-c: A multilingual corpus for end-to-end speech translation}.
\newblock \emph{Computer Speech \& Language}, 66:101--155.

\bibitem[{Cho et~al.(2019)Cho, Kim, Kim, and Kim}]{cho-etal-2019-measuring}
Won~Ik Cho, Ji~Won Kim, Seok~Min Kim, and Nam~Soo Kim. 2019.
\newblock \href {https://doi.org/10.18653/v1/W19-3824} {On measuring gender
  bias in translation of gender-neutral pronouns}.
\newblock In \emph{Proceedings of the First Workshop on Gender Bias in Natural
  Language Processing}, pages 173--181, Florence, Italy. Association for
  Computational Linguistics.

\bibitem[{Chorowski and Jaitly(2017)}]{Chorowski2016TowardsBD}
Jan Chorowski and Navdeep Jaitly. 2017.
\newblock \href
  {http://www.isca-speech.org/archive/Interspeech\_2017/abstracts/0343.html}
  {Towards better decoding and language model integration in sequence to
  sequence models}.
\newblock In \emph{Interspeech 2017, 18th Annual Conference of the
  International Speech Communication Association, Stockholm, Sweden, August
  20-24, 2017}, pages 523--527. International Speech Communication Association.

\bibitem[{Corbett(2013)}]{corbett}
Greville~G. Corbett. 2013.
\newblock \emph{The Expression of Gender}.
\newblock De Gruyter.

\bibitem[{Costa{-}juss{\`{a}} et~al.(2022)Costa{-}juss{\`{a}}, Basta, and
  G{\'{a}}llego}]{CostaJussa-GenderArchitecture}
Marta~R. Costa{-}juss{\`{a}}, Christine Basta, and Gerard~I. G{\'{a}}llego.
  2022.
\newblock \href {https://aclanthology.org/2022.lrec-1.230} {Evaluating gender
  bias in speech translation}.
\newblock In \emph{Proceedings of the Language Resources and Evaluation
  Conference}, pages 2141--2147, Marseille, France. European Language Resources
  Association.

\bibitem[{Crawford(2017)}]{crawford2017trouble}
Kate Crawford. 2017.
\newblock \href {https://www.youtube.com/watch?v=fMym_BKWQzk} {{The Trouble
  with Bias}}.
\newblock In \emph{Conference on Neural Information Processing Systems (NIPS)
  -- Keynote}, Long Beach, California, USA.

\bibitem[{Di~Gangi et~al.(2020)Di~Gangi, Gaido, Negri, and
  Turchi}]{di-gangi-etal-2020-target}
Mattia~A. Di~Gangi, Marco Gaido, Matteo Negri, and Marco Turchi. 2020.
\newblock \href {https://aclanthology.org/2020.amta-research.13} {On target
  segmentation for direct speech translation}.
\newblock In \emph{Proceedings of the 14th Conference of the Association for
  Machine Translation in the Americas (Volume 1: Research Track)}, pages
  137--150, Virtual. Association for Machine Translation in the Americas.

\bibitem[{Gaido et~al.(2021)Gaido, Cettolo, Negri, and
  Turchi}]{gaido-etal-2021-ctc}
Marco Gaido, Mauro Cettolo, Matteo Negri, and Marco Turchi. 2021.
\newblock \href {https://doi.org/10.18653/v1/2021.eacl-main.57} {{CTC}-based
  compression for direct speech translation}.
\newblock In \emph{Proceedings of the 16th Conference of the European Chapter
  of the Association for Computational Linguistics: Main Volume}, pages
  690--696, Online. Association for Computational Linguistics.

\bibitem[{Gaido et~al.(2022)Gaido, Papi, Fucci, Fiameni, Negri, and
  Turchi}]{gaido-etal-2022-efficient-yet}
Marco Gaido, Sara Papi, Dennis Fucci, Giuseppe Fiameni, Matteo Negri, and Marco
  Turchi. 2022.
\newblock \href {https://doi.org/10.18653/v1/2022.iwslt-1.13} {Efficient yet
  competitive speech translation: {FBK}@{IWSLT}2022}.
\newblock In \emph{Proceedings of the 19th International Conference on Spoken
  Language Translation (IWSLT 2022)}, pages 177--189, Dublin, Ireland.
  Association for Computational Linguistics.

\bibitem[{Gaido et~al.(2020)Gaido, Savoldi, Bentivogli, Negri, and
  Turchi}]{gaido-etal-2020-breeding}
Marco Gaido, Beatrice Savoldi, Luisa Bentivogli, Matteo Negri, and Marco
  Turchi. 2020.
\newblock \href {https://doi.org/10.18653/v1/2020.coling-main.350} {Breeding
  gender-aware direct speech translation systems}.
\newblock In \emph{Proceedings of the 28th International Conference on
  Computational Linguistics}, pages 3951--3964, Barcelona, Spain (Online).
  International Committee on Computational Linguistics.

\bibitem[{Gulati et~al.(2020)Gulati, Qin, Chiu, Parmar, Zhang, Yu, Han, Wang,
  Zhang, Wu, and Pang}]{gulati20_interspeech}
Anmol Gulati, James Qin, Chung-Cheng Chiu, Niki Parmar, Yu~Zhang, Jiahui Yu,
  Wei Han, Shibo Wang, Zhengdong Zhang, Yonghui Wu, and Ruoming Pang. 2020.
\newblock \href {https://doi.org/10.21437/Interspeech.2020-3015} {{Conformer:
  Convolution-augmented Transformer for Speech Recognition}}.
\newblock In \emph{Proceedings of the 21st Annual Conference of the
  International Speech Communication Association}, pages 5036--5040, Shanghai,
  China (Online). International Speech Communication Association.

\bibitem[{G{\"{u}}l{\c{c}}ehre et~al.(2015)G{\"{u}}l{\c{c}}ehre, Firat, Xu,
  Cho, Barrault, Lin, Bougares, Schwenk, and Bengio}]{gulchere2015}
{\c{C}}aglar G{\"{u}}l{\c{c}}ehre, Orhan Firat, Kelvin Xu, Kyunghyun Cho,
  Lo{\"{\i}}c Barrault, Huei{-}Chi Lin, Fethi Bougares, Holger Schwenk, and
  Yoshua Bengio. 2015.
\newblock \href {http://arxiv.org/abs/1503.03535} {On using monolingual corpora
  in neural machine translation}.
\newblock \emph{CoRR}, abs/1503.03535.

\bibitem[{G{\"u}lçehre et~al.(2017)G{\"u}lçehre, Firat, Xu, Cho, and
  Bengio}]{gulchere2017}
{\c{C}}aglar G{\"u}lçehre, Orhan Firat, Kelvin Xu, Kyunghyun Cho, and Yoshua
  Bengio. 2017.
\newblock \href {https://doi.org/https://doi.org/10.1016/j.csl.2017.01.014} {On
  integrating a language model into neural machine translation}.
\newblock \emph{Computer Speech \& Language}, 45:137--148.

\bibitem[{Gygax et~al.(2019)Gygax, Elmiger, Zufferey, Garnham, Sczesny, von
  Stockhausen, and Oakhill}]{gygax}
Pascal~M. Gygax, Daniel Elmiger, Sandrine Zufferey, Alan Garnham, Sabine
  Sczesny, Friederike von Stockhausen, Lisa~Braun, and Jane Oakhill. 2019.
\newblock \href {https://doi.org/10.3389/fpsyg.2019.01604} {A language index of
  grammatical gender dimensions to study the impact of grammatical gender on
  the way we perceive women and men}.
\newblock \emph{Frontiers in Psychology}, 10.

\bibitem[{Hori et~al.(2017)Hori, Watanabe, and Hershey}]{Hori2017}
Takaaki Hori, Shinji Watanabe, and John~R. Hershey. 2017.
\newblock \href {https://doi.org/10.1109/ASRU.2017.8268948} {Multi-level
  language modeling and decoding for open vocabulary end-to-end speech
  recognition}.
\newblock In \emph{2017 IEEE Automatic Speech Recognition and Understanding
  Workshop (ASRU)}, pages 287--293. Institute of Electrical and Electronics
  Engineers.

\bibitem[{Huang et~al.(2022)Huang, Peyser, Sainath, Pang, Strohman, and
  Kumar}]{Huang2022SentenceSelectLL}
W.~Ronny Huang, Cal Peyser, Tara~N. Sainath, Ruoming Pang, Trevor~D. Strohman,
  and Shankar Kumar. 2022.
\newblock \href {https://doi.org/10.21437/Interspeech.2022-10820}
  {Sentence-select: Large-scale language model data selection for rare-word
  speech recognition}.
\newblock In \emph{Interspeech 2022, 23rd Annual Conference of the
  International Speech Communication Association, Incheon, Korea, 18-22
  September 2022}, pages 689--693. International Speech Communication
  Association.

\bibitem[{Inaguma et~al.(2019)Inaguma, Cho, Baskar, Kawahara, and
  Watanabe}]{Inaguma_2019_transfer}
Hirofumi Inaguma, Jaejin Cho, Murali~Karthick Baskar, Tatsuya Kawahara, and
  Shinji Watanabe. 2019.
\newblock \href {https://doi.org/10.1109/ICASSP.2019.8682918} {Transfer
  learning of language-independent end-to-end asr with language model fusion}.
\newblock In \emph{ICASSP 2019 - 2019 IEEE International Conference on
  Acoustics, Speech and Signal Processing (ICASSP)}, pages 6096--6100.
  Institute of Electrical and Electronics Engineers.

\bibitem[{Inaguma et~al.(2021)Inaguma, Kawahara, and
  Watanabe}]{inaguma-etal-2021-source}
Hirofumi Inaguma, Tatsuya Kawahara, and Shinji Watanabe. 2021.
\newblock \href {https://doi.org/10.18653/v1/2021.naacl-main.150} {Source and
  target bidirectional knowledge distillation for end-to-end speech
  translation}.
\newblock In \emph{Proceedings of the 2021 Conference of the North American
  Chapter of the Association for Computational Linguistics: Human Language
  Technologies}, pages 1872--1881, Online. Association for Computational
  Linguistics.

\bibitem[{Irie et~al.(2019)Irie, Zeyer, Schl{\"{u}}ter, and
  Ney}]{irie19b_interspeech}
Kazuki Irie, Albert Zeyer, Ralf Schl{\"{u}}ter, and Hermann Ney. 2019.
\newblock \href {https://doi.org/10.21437/Interspeech.2019-2225} {Language
  modeling with deep transformers}.
\newblock In \emph{Interspeech 2019, 20th Annual Conference of the
  International Speech Communication Association, Graz, Austria, 15-19
  September 2019}, pages 3905--3909. International Speech Communication
  Association.

\bibitem[{Kannan et~al.(2018)Kannan, Wu, Nguyen, Sainath, Chen, and
  Prabhavalkar}]{Kannan2018}
Anjuli Kannan, Yonghui Wu, Patrick Nguyen, Tara~N. Sainath, ZhiJeng Chen, and
  Rohit Prabhavalkar. 2018.
\newblock \href {https://doi.org/10.1109/ICASSP.2018.8462682} {An analysis of
  incorporating an external language model into a sequence-to-sequence model}.
\newblock In \emph{2018 IEEE International Conference on Acoustics, Speech and
  Signal Processing (ICASSP)}, pages 5824--5828. Institute of Electrical and
  Electronics Engineers.

\bibitem[{Kingma and Ba(2015)}]{DBLP:journals/corr/KingmaB14}
Diederik~P. Kingma and Jimmy Ba. 2015.
\newblock \href {http://arxiv.org/abs/1412.6980} {Adam: {A} method for
  stochastic optimization}.
\newblock In \emph{Proceedings of the 3rd International Conference on Learning
  Representations}, San Diego, USA.

\bibitem[{Koehn(2004)}]{koehn-2004-statistical}
Philipp Koehn. 2004.
\newblock \href {https://aclanthology.org/W04-3250} {Statistical significance
  tests for machine translation evaluation}.
\newblock In \emph{Proceedings of the 2004 Conference on Empirical Methods in
  Natural Language Processing}, pages 388--395, Barcelona, Spain. Association
  for Computational Linguistics.

\bibitem[{Kudo and Richardson(2018)}]{kudo-richardson-2018-sentencepiece}
Taku Kudo and John Richardson. 2018.
\newblock \href {https://doi.org/10.18653/v1/D18-2012} {{S}entence{P}iece: A
  simple and language independent subword tokenizer and detokenizer for neural
  text processing}.
\newblock In \emph{Proceedings of the 2018 Conference on Empirical Methods in
  Natural Language Processing: System Demonstrations}, pages 66--71, Brussels,
  Belgium. Association for Computational Linguistics.

\bibitem[{Lauscher et~al.(2022)Lauscher, Crowley, and Hovy}]{hovy}
Anne Lauscher, Archie Crowley, and Dirk Hovy. 2022.
\newblock \href {https://doi.org/10.48550/ARXIV.2202.11923} {Welcome to the
  modern world of pronouns: Identity-inclusive natural language processing
  beyond gender}.
\newblock \emph{CoRR}, abs/2202.11923.

\bibitem[{Liu et~al.(2020)Liu, Zhu, Zhang, and Zong}]{liu2020bridging}
Yuchen Liu, Junnan Zhu, Jiajun Zhang, and Chengqing Zong. 2020.
\newblock \href {http://arxiv.org/abs/2010.14920} {Bridging the modality gap
  for speech-to-text translation}.
\newblock \emph{CoRR}, abs/2010.14920.

\bibitem[{Liu et~al.(2022)Liu, Ma, Xu, He, Ma, and Zhang}]{liu22j_interspeech}
Yufei Liu, Rao Ma, Haihua Xu, Yi~He, Zejun Ma, and Weibin Zhang. 2022.
\newblock \href {https://doi.org/10.21437/Interspeech.2022-606} {Internal
  language model estimation through explicit context vector learning for
  attention-based encoder-decoder {ASR}}.
\newblock In \emph{Interspeech 2022, 23rd Annual Conference of the
  International Speech Communication Association, Incheon, Korea, 18-22
  September 2022}, pages 1666--1670. International Speech Communication
  Association.

\bibitem[{Matar et~al.(2016)Matar, Portes, Lancia, Legou, and Baider}]{matar}
Nayla Matar, Cristel Portes, Leonardo Lancia, Thierry Legou, and Fabienne
  Baider. 2016.
\newblock \href {https://hal.archives-ouvertes.fr/hal-01459619} {{Voice quality
  and gender stereotypes: A study on Lebanese women with Reinke's edema}}.
\newblock \emph{{Journal of Speech, Language, and Hearing Research}},
  59(6):1608--1617.

\bibitem[{McDermott et~al.(2019)McDermott, Sak, and Variani}]{McDermott2019ADR}
Erik McDermott, Hasim Sak, and Ehsan Variani. 2019.
\newblock \href {https://doi.org/10.1109/ASRU46091.2019.9003790.} {A density
  ratio approach to language model fusion in end-to-end automatic speech
  recognition}.
\newblock In \emph{Proceedings of 2019 IEEE Automatic Speech Recognition and
  Understanding Workshop}, pages 434--441, Sentosa, Singapore. Institute of
  Electrical and Electronics Engineers.

\bibitem[{Menezes et~al.(2022)Menezes, {de Lira}, de~Araújo, {de Almeida},
  de~Oliveira Camargo~Gomes, Moraes, and Lucena}]{MENEZES2022}
Danielle~Pereira Menezes, Zulina~Souza {de Lira}, Ana Nery~Barbosa de~Araújo,
  Anna Alice~Figueirêdo {de Almeida}, Adriana de~Oliveira Camargo~Gomes,
  Bruno~Teixeira Moraes, and Jonia~Alves Lucena. 2022.
\newblock \href {https://doi.org/https://doi.org/10.1016/j.jvoice.2021.12.020}
  {Prosodic differences in the voices of transgender and cisgender women:
  Self-perception of voice - an auditory and acoustic analysis}.
\newblock \emph{Journal of Voice}.

\bibitem[{Meng et~al.(2021{\natexlab{a}})Meng, Kanda, Gaur, Parthasarathy, Sun,
  Lu, Chen, Li, and Gong}]{Meng-ICASSP}
Zhong Meng, Naoyuki Kanda, Yashesh Gaur, Sarangarajan Parthasarathy, Eric Sun,
  Liang Lu, Xie Chen, Jinyu Li, and Yifan Gong. 2021{\natexlab{a}}.
\newblock \href {https://doi.org/10.1109/ICASSP39728.2021.9415039} {Internal
  language model training for domain-adaptive end-to-end speech recognition}.
\newblock In \emph{Proceedings of 2021 IEEE International Conference on
  Acoustics, Speech and Signal Processing}, pages 7338--7342, Toronto, Canada
  (Online). Institute of Electrical and Electronics Engineers.

\bibitem[{Meng et~al.(2021{\natexlab{b}})Meng, Parthasarathy, Sun, Gaur, Kanda,
  Lu, Chen, Zhao, Li, and Gong}]{Meng-SLT}
Zhong Meng, Sarangarajan Parthasarathy, Eric Sun, Yashesh Gaur, Naoyuki Kanda,
  Liang Lu, Xie Chen, Rui Zhao, Jinyu Li, and Yifan Gong. 2021{\natexlab{b}}.
\newblock \href {https://doi.org/10.1109/SLT48900.2021.9383515} {Internal
  language model estimation for domain-adaptive end-to-end speech recognition}.
\newblock In \emph{Proceedings of 2021 IEEE Spoken Language Technology
  Workshop}, pages 243--250, Shenzhen, China (Online). Institute of Electrical
  and Electronics Engineers.

\bibitem[{Meng et~al.(2023)Meng, Wang, Prabhavalkar, Sainath, Chen, Variani,
  Zhang, Li, Rosenberg, and Ramabhadran}]{10095249}
Zhong Meng, Weiran Wang, Rohit Prabhavalkar, Tara~N. Sainath, Tongzhou Chen,
  Ehsan Variani, Yu~Zhang, Bo~Li, Andrew Rosenberg, and Bhuvana Ramabhadran.
  2023.
\newblock \href {https://doi.org/10.1109/ICASSP49357.2023.10095249} {Jeit:
  Joint end-to-end model and internal language model training for speech
  recognition}.
\newblock In \emph{ICASSP 2023 - 2023 IEEE International Conference on
  Acoustics, Speech and Signal Processing (ICASSP)}, pages 1--5. Institute of
  Electrical and Electronics Engineers.

\bibitem[{Meng et~al.(2021{\natexlab{c}})Meng, Wu, Kanda, Lu, Chen, Ye, Sun,
  Li, and Gong}]{Meng-Interspeech}
Zhong Meng, Yu~Wu, Naoyuki Kanda, Liang Lu, Xie Chen, Guoli Ye, Eric Sun, Jinyu
  Li, and Yifan Gong. 2021{\natexlab{c}}.
\newblock \href {https://doi.org/10.21437/Interspeech.2021-2075} {Minimum word
  error rate training with language model fusion for end-to-end speech
  recognition}.
\newblock In \emph{Proceedings of the 22nd Annual Conference of the
  International Speech Communication Association}, pages 2596--2600, Brno,
  Czechia. International Speech Communication Association.

\bibitem[{Mulac et~al.(2001)Mulac, Bradac, and Gibbons}]{Mulac2001}
Anthony Mulac, James~J. Bradac, and Pamela Gibbons. 2001.
\newblock \href {https://doi.org/10.1111/j.1468-2958.2001.tb00778.x} {Empirical
  {S}upport for the {G}ender‐{a}s‐{C}ulture {H}ypothesis}.
\newblock \emph{Human Communication Research}, 27:121--152.

\bibitem[{Ott et~al.(2019)Ott, Edunov, Baevski, Fan, Gross, Ng, Grangier, and
  Auli}]{ott-etal-2019-fairseq}
Myle Ott, Sergey Edunov, Alexei Baevski, Angela Fan, Sam Gross, Nathan Ng,
  David Grangier, and Michael Auli. 2019.
\newblock \href {https://doi.org/10.18653/v1/N19-4009} {fairseq: A fast,
  extensible toolkit for sequence modeling}.
\newblock In \emph{Proceedings of the 2019 Conference of the North {A}merican
  Chapter of the Association for Computational Linguistics (Demonstrations)},
  pages 48--53, Minneapolis, Minnesota. Association for Computational
  Linguistics.

\bibitem[{Pereira et~al.(2018)Pereira, Dassie-Leite, Pereira, Cavichiolo, Rosa,
  and Fugmann}]{pereira}
Amanda~Maria Pereira, Ana~Paula Dassie-Leite, Eliane~Cristina Pereira,
  Juliana~Benthien Cavichiolo, Marcelo de~Oliveira Rosa, and Elmar~Allen
  Fugmann. 2018.
\newblock \href {https://doi.org/10.1590/2317-1782/20182017046} {Auditory
  perception of lay judges about gender identification of women with reinke's
  edema}.
\newblock \emph{CoDAS}, 30(4).

\bibitem[{Post(2018)}]{post-2018-call}
Matt Post. 2018.
\newblock \href {https://doi.org/10.18653/v1/W18-6319} {A call for clarity in
  reporting {BLEU} scores}.
\newblock In \emph{Proceedings of the Third Conference on Machine Translation:
  Research Papers}, pages 186--191, Brussels, Belgium. Association for
  Computational Linguistics.

\bibitem[{Prates et~al.(2020)Prates, Avelar, and Lamb}]{Prates2018AssessingGB}
Marcelo O.~R. Prates, Pedro H.~C. Avelar, and Lu{\'i}s~C. Lamb. 2020.
\newblock \href {https://doi.org/10.1007/s00521-019-04144-6} {Assessing gender
  bias in machine translation: a case study with google translate}.
\newblock \emph{Neural Computing and Applications}, 32:6363–6381.

\bibitem[{Sainath et~al.(2021)Sainath, He, Narayanan, Botros, Pang, Rybach,
  Allauzen, Variani, Qin, Le{-}The, Chang, Li, Gulati, Yu, Chiu, Caseiro, Li,
  Liang, and Rondon}]{Sainath2021AnES}
Tara~N. Sainath, Yanzhang He, Arun Narayanan, Rami Botros, Ruoming Pang, David
  Rybach, Cyril Allauzen, Ehsan Variani, James Qin, Quoc{-}Nam Le{-}The,
  Shuo{-}Yiin Chang, Bo~Li, Anmol Gulati, Jiahui Yu, Chung{-}Cheng Chiu,
  Diamantino Caseiro, Wei Li, Qiao Liang, and Pat Rondon. 2021.
\newblock \href {https://doi.org/10.21437/Interspeech.2021-206} {An efficient
  streaming non-recurrent on-device end-to-end model with improvements to
  rare-word modeling}.
\newblock In \emph{Interspeech 2021, 22nd Annual Conference of the
  International Speech Communication Association, Brno, Czechia, 30 August - 3
  September 2021}, pages 1777--1781. International Speech Communication
  Association.

\bibitem[{Salesky et~al.(2019)Salesky, Sperber, and
  Black}]{salesky-etal-2019-exploring}
Elizabeth Salesky, Matthias Sperber, and Alan~W Black. 2019.
\newblock \href {https://doi.org/10.18653/v1/P19-1179} {Exploring phoneme-level
  speech representations for end-to-end speech translation}.
\newblock In \emph{Proceedings of the 57th Annual Meeting of the Association
  for Computational Linguistics}, pages 1835--1841, Florence, Italy.
  Association for Computational Linguistics.

\bibitem[{Savoldi et~al.(2021)Savoldi, Gaido, Bentivogli, Negri, and
  Turchi}]{savoldi-etal-2021-gender}
Beatrice Savoldi, Marco Gaido, Luisa Bentivogli, Matteo Negri, and Marco
  Turchi. 2021.
\newblock \href {https://doi.org/10.1162/tacl_a_00401} {Gender bias in machine
  translation}.
\newblock \emph{Transactions of the Association for Computational Linguistics},
  9:845--874.

\bibitem[{Sennrich et~al.(2016)Sennrich, Haddow, and
  Birch}]{sennrich-etal-2016-neural}
Rico Sennrich, Barry Haddow, and Alexandra Birch. 2016.
\newblock \href {https://doi.org/10.18653/v1/P16-1162} {Neural machine
  translation of rare words with subword units}.
\newblock In \emph{Proceedings of the 54th Annual Meeting of the Association
  for Computational Linguistics (Volume 1: Long Papers)}, pages 1715--1725,
  Berlin, Germany. Association for Computational Linguistics.

\bibitem[{Shan et~al.(2019)Shan, Weng, Wang, Su, Luo, Yu, and Xie}]{Shan}
Changhao Shan, Chao Weng, Guangsen Wang, Dan Su, Min Luo, Dong Yu, and Lei Xie.
  2019.
\newblock \href {https://doi.org/10.1109/ICASSP.2019.8682490} {Component
  fusion: Learning replaceable language model component for end-to-end speech
  recognition system}.
\newblock In \emph{Proceedings of 2019 IEEE International Conference on
  Acoustics, Speech and Signal Processing}, pages 5361--5635, Brighton, United
  Kingdom. Institute of Electrical and Electronics Engineers.

\bibitem[{Sriram et~al.(2018)Sriram, Jun, Satheesh, and Coates}]{sriram}
Anuroop Sriram, Heewoo Jun, Sanjeev Satheesh, and Adam Coates. 2018.
\newblock \href {https://doi.org/10.21437/Interspeech.2018-1392} {{Cold Fusion:
  Training Seq2Seq Models Together with Language Models}}.
\newblock In \emph{Proceedings of the 19th Annual Conference of the
  International Speech Communication Association}, pages 387--391, Hyderabad,
  India. International Speech Communication Association.

\bibitem[{Stahlberg et~al.(2007)Stahlberg, Braun, Irmen, and
  Sczesny}]{Stahlberg-gender}
Dagmar Stahlberg, Friederike Braun, Lisa Irmen, and Sabine Sczesny. 2007.
\newblock \href {https://psycnet.apa.org/record/2007-01308-006} {Representation
  of the sexes in language}.
\newblock In Klaus Fiedler, editor, \emph{Social Communication}, pages
  163--187. Psychology Press.

\bibitem[{Stahlberg et~al.(2018)Stahlberg, Cross, and Stoyanov}]{stahlberg}
Felix Stahlberg, James Cross, and Veselin Stoyanov. 2018.
\newblock \href {https://doi.org/10.18653/v1/W18-6321} {Simple fusion: Return
  of the language model}.
\newblock In \emph{Proceedings of the Third Conference on Machine Translation:
  Research Papers}, pages 204--211, Brussels, Belgium. Association for
  Computational Linguistics.

\bibitem[{Tatman(2017)}]{tatman-2017-gender}
Rachael Tatman. 2017.
\newblock \href {https://doi.org/10.18653/v1/W17-1606} {Gender and dialect bias
  in {Y}ou{T}ube{'}s automatic captions}.
\newblock In \emph{Proceedings of the First {ACL} Workshop on Ethics in Natural
  Language Processing}, pages 53--59, Valencia, Spain. Association for
  Computational Linguistics.

\bibitem[{Variani et~al.(2020)Variani, Rybach, Allauzen, and
  Riley}]{VarianiHAT}
Ehsan Variani, David Rybach, Cyril Allauzen, and Michael Riley. 2020.
\newblock \href {https://doi.org/10.1109/ICASSP40776.2020.9053600} {Hybrid
  autoregressive transducer (hat)}.
\newblock In \emph{ICASSP 2020 - 2020 IEEE International Conference on
  Acoustics, Speech and Signal Processing (ICASSP)}, pages 6139--6143.
  Institute of Electrical and Electronics Engineers.

\bibitem[{Vaswani et~al.(2017)Vaswani, Shazeer, Parmar, Uszkoreit, Jones,
  Gomez, Kaiser, and Polosukhin}]{transformer}
Ashish Vaswani, Noam Shazeer, Niki Parmar, Jakob Uszkoreit, Llion Jones,
  Aidan~N Gomez, Łukasz Kaiser, and Illia Polosukhin. 2017.
\newblock \href
  {https://proceedings.neurips.cc/paper/2017/file/3f5ee243547dee91fbd053c1c4a845aa-Paper.pdf}
  {Attention is all you need}.
\newblock In \emph{Proceedings of the 31st International Conference on Neural
  Information Processing Systems}, Long Beach, USA. Curran Associates Inc.

\bibitem[{Villas-Bôas et~al.(2021)Villas-Bôas, Schwarz, Fontanari, Costa,
  Cardoso~da Silva, Schneider, Cielo, Spritzer, and
  Rodrigues~Lobato}]{VillasBoas}
Anna~Paula Villas-Bôas, Karine Schwarz, Anna Martha~Vaitses Fontanari,
  Angelo~Brandelli Costa, Dhiordan Cardoso~da Silva, Maiko~Abel Schneider,
  Carla~Aparecida Cielo, Poli~Mara Spritzer, and Maria~Inês Rodrigues~Lobato.
  2021.
\newblock \href {https://doi.org/10.3389/fpsyg.2021.622526} {Acoustic measures
  of brazilian transgender women's voices: A case–control study}.
\newblock \emph{Frontiers in Psychology}, 12.

\bibitem[{Wang et~al.(2020)Wang, Tang, Ma, Wu, Okhonko, and
  Pino}]{wang2020fairseqs2t}
Changhan Wang, Yun Tang, Xutai Ma, Anne Wu, Dmytro Okhonko, and Juan Pino.
  2020.
\newblock \href {https://aclanthology.org/2020.aacl-demo.6} {Fairseq {S}2{T}:
  Fast speech-to-text modeling with fairseq}.
\newblock In \emph{Proceedings of the 1st Conference of the Asia-Pacific
  Chapter of the Association for Computational Linguistics and the 10th
  International Joint Conference on Natural Language Processing: System
  Demonstrations}, pages 33--39, Suzhou, China. Association for Computational
  Linguistics.

\bibitem[{Zeineldeen et~al.(2021)Zeineldeen, Glushko, Michel, Zeyer,
  Schl{\"{u}}ter, and Ney}]{Zeineldeen}
Mohammad Zeineldeen, Aleksandr Glushko, Wilfried Michel, Albert Zeyer, Ralf
  Schl{\"{u}}ter, and Hermann Ney. 2021.
\newblock \href {https://doi.org/10.21437/Interspeech.2021-1255} {Investigating
  methods to improve language model integration for attention-based
  encoder-decoder {ASR} models}.
\newblock In \emph{Proceedings of the 22nd Annual Conference of the
  International Speech Communication Association}, pages 2856--2860, Brno,
  Czechia. International Speech Communication Association.

\end{thebibliography}

\appendix

\begin{table*}[!htb]
\centering
\small
\setlength{\tabcolsep}{4pt}
\begin{tabular}{l||rrrr||rrrr||rrrr}
\hline
\multirow{3}{*}{\textbf{Models}} & \multicolumn{4}{|c||}{\textbf{en-es}}   & \multicolumn{4}{c||}{\textbf{en-fr}}   & \multicolumn{4}{c}{\textbf{en-it}} \\ 
\cline{2-13} 
& \multicolumn{2}{c|}{M}  & \multicolumn{2}{c||}{F} & \multicolumn{2}{c|}{M} & \multicolumn{2}{c||}{F}  & \multicolumn{2}{c|}{M} & \multicolumn{2}{c}{F} \\
\cline{2-13} 
& \multicolumn{1}{c}{$\beta_{ILM}$}    & \multicolumn{1}{c|}{$\beta_{ELM}$}    & \multicolumn{1}{c}{$\beta_{ILM}$}    & \multicolumn{1}{c||}{$\beta_{ELM}$} & \multicolumn{1}{c}{$\beta_{ILM}$}    & \multicolumn{1}{c|}{$\beta_{ELM}$}    & \multicolumn{1}{c}{$\beta_{ILM}$}    & \multicolumn{1}{c||}{$\beta_{ELM}$} & \multicolumn{1}{c}{$\beta_{ILM}$}    & \multicolumn{1}{c|}{$\beta_{ELM}$}    & \multicolumn{1}{c}{$\beta_{ILM}$}    & \multicolumn{1}{c}{$\beta_{ELM}$} \\ \hline
M\textsubscript{B-ILM+ELM} & \multicolumn{1}{r}{0.200} & 0.250 & \multicolumn{1}{|r}{0.285} & \multicolumn{1}{r||}{0.390} & \multicolumn{1}{r}{0.155} & 0.245 & \multicolumn{1}{|r}{0.215} & 0.355 & \multicolumn{1}{r}{0.125} & 0.310 & \multicolumn{1}{|r}{0.195} & 0.305 \\
M\textsubscript{B+ELM} & \multicolumn{1}{r}{-} & \multicolumn{1}{r|}{0.145} & \multicolumn{1}{r}{-} & 0.310 & \multicolumn{1}{r}{-} & \multicolumn{1}{r|}{0.235} & \multicolumn{1}{r}{-} & 0.300 & \multicolumn{1}{r}{-} & \multicolumn{1}{r|}{0.195} & \multicolumn{1}{r}{-} & 0.275 \\
\hline
\end{tabular}
\caption{Mean of the 
optimal values for $\beta_{ILM}$ and $\beta_{ELM}$ found using 10-fold cross-validation.}
\label{tab:mean-beta}
\end{table*}

\begin{figure*}[!htb]
     \centering
     \begin{subfigure}[b]{0.272\textwidth}
         \centering
         \includegraphics[width=\textwidth]{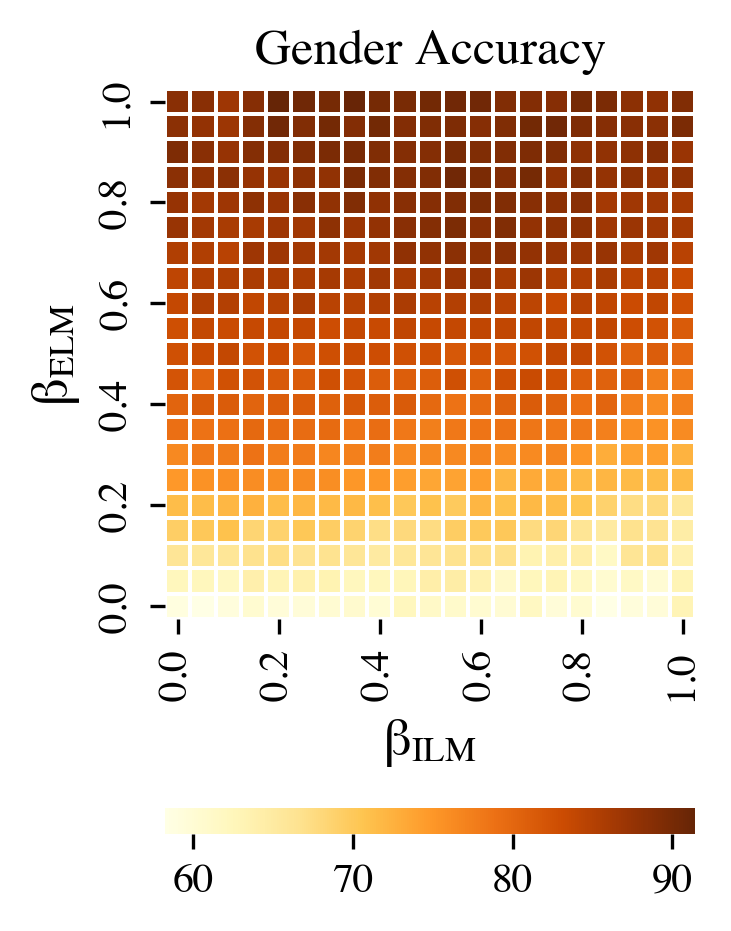}
         \includegraphics[width=\textwidth]{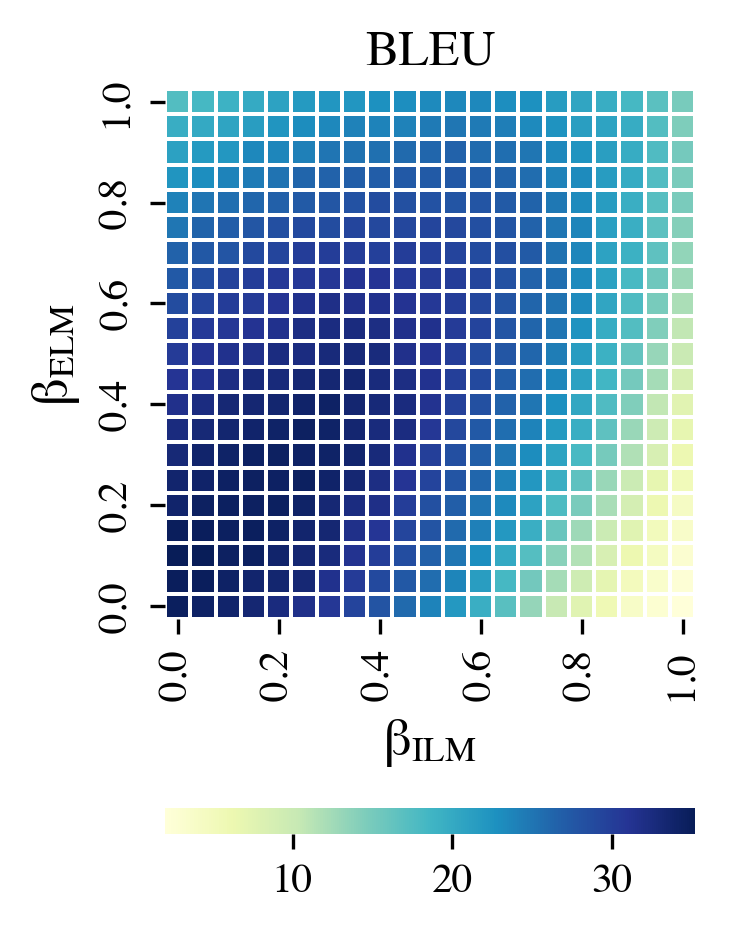}
         \caption{en-es}
     \end{subfigure}
     \begin{subfigure}[b]{0.272\textwidth}
         \centering
         \includegraphics[width=\textwidth]{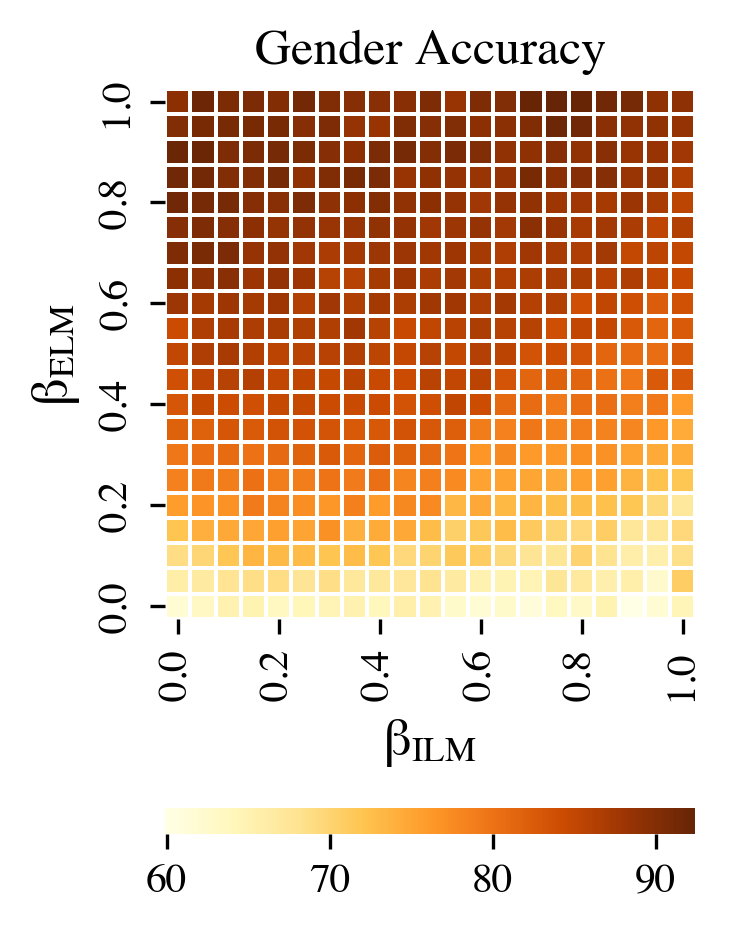}
         \includegraphics[width=\textwidth]{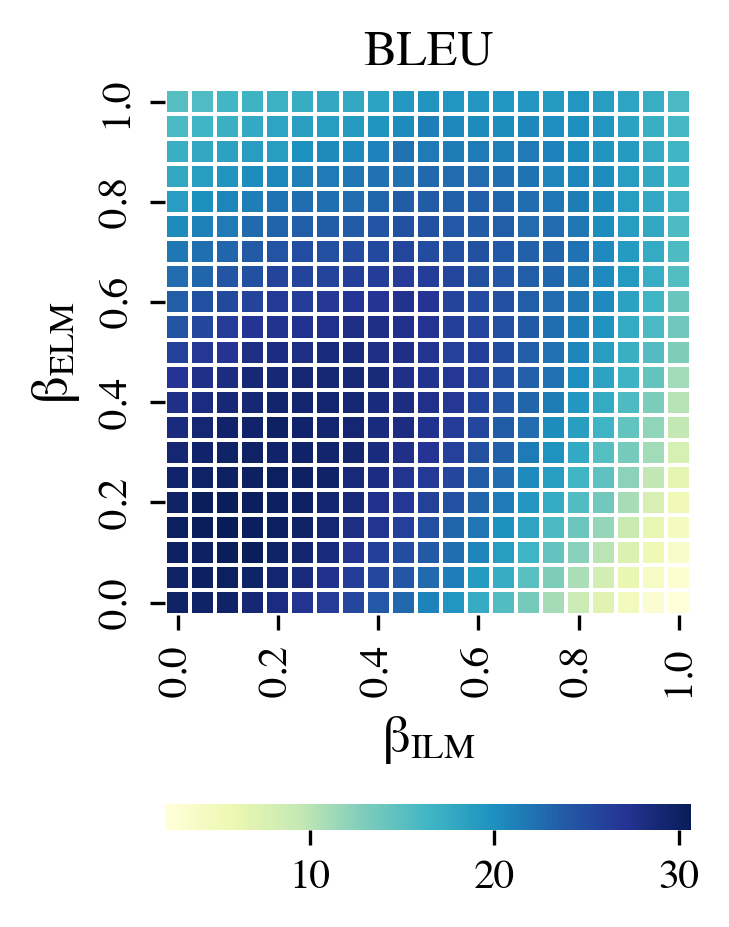}
         \caption{en-fr}
     \end{subfigure}
     \begin{subfigure}[b]{0.272\textwidth}
         \centering
         \includegraphics[width=\textwidth]{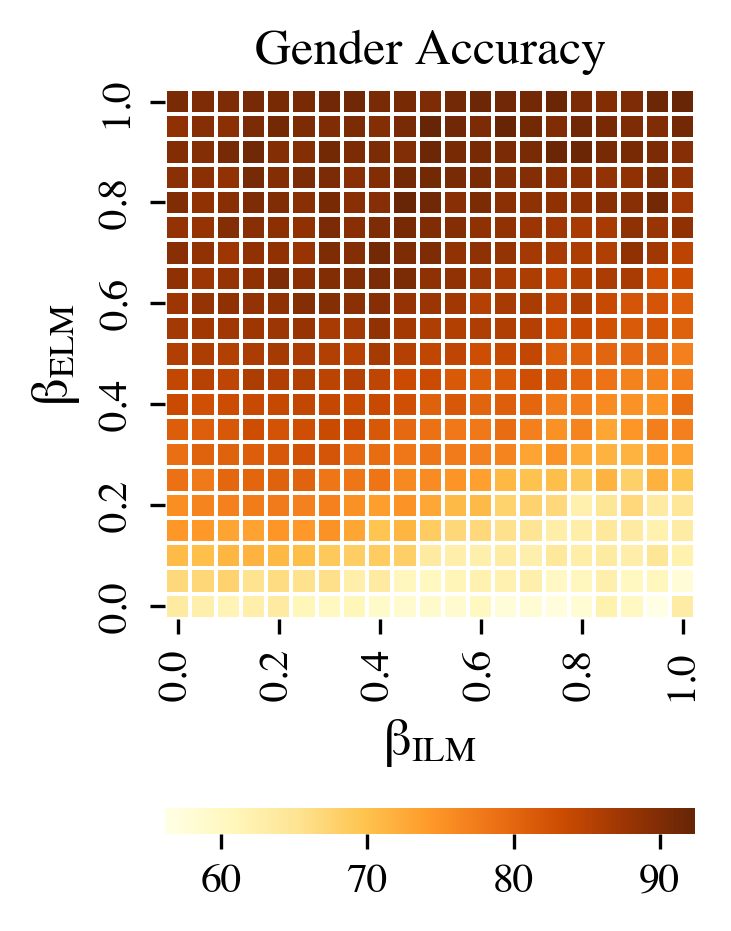}
         \includegraphics[width=\textwidth]{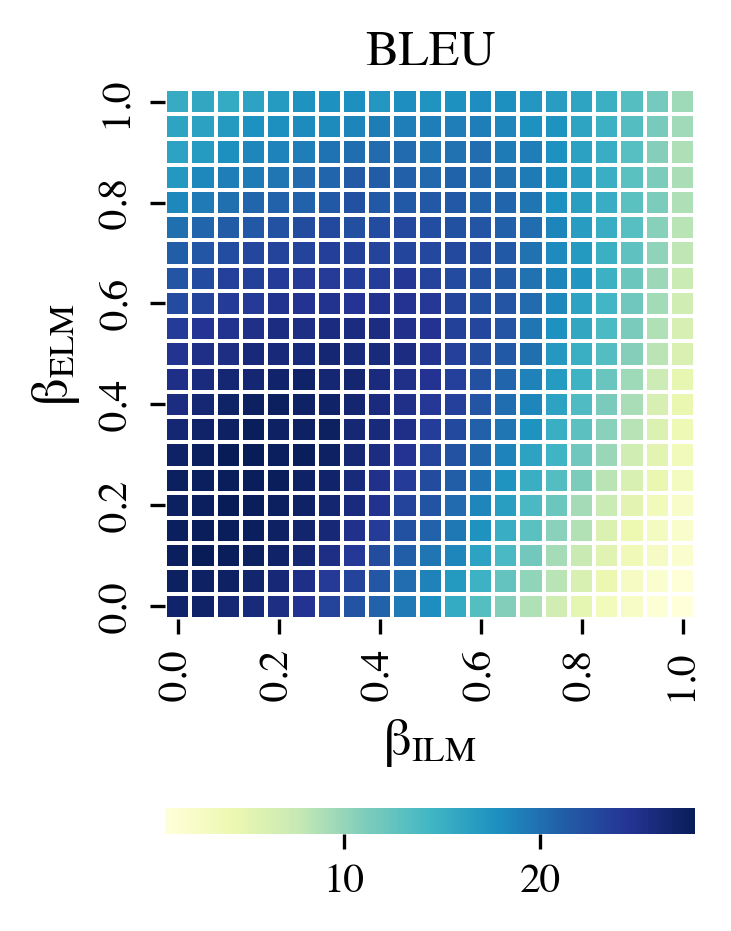}
         \caption{en-it}
     \end{subfigure}
        \caption{
    BLEU and gender accuracy heatmaps
    with different combinations of 
    $\beta_{ILM}$ and $\beta_{ELM}$ for all language pairs.}
        \label{fig:heatmap}
\end{figure*}

\section{Contributions of $\beta_{ILM}$-$\beta_{ELM}$}
\label{app:beta}

As stated in \S\ref{sec:method}, our method relies on two hyper-parameters 
($\beta_{ELM}$ and $\beta_{ILM}$).
In this section, we report their optimal values (\S\ref{app:optimal_combinations}), and discuss the impact of varying these values on the results (\S\ref{app:heatmaps}).

\subsection{Optimal $\beta_{ILM}$-$\beta_{ELM}$ Combinations}
\label{app:optimal_combinations}

In the lack of a validation set with the same characteristics of MuST-SHE, we used this same benchmark for a 10-fold cross validation.
At each iteration, we translate the held-out data with the pair $(\beta_{ILM}, \beta_{ELM}) \in \{0.00,0.05,\ldots 0.95,1.00 \}^2$ that maximizes the harmonic mean between gender accuracy and BLEU (see \S\ref{sec:method}) on the validation folds.
At the end of this process, the whole MuST-SHE was fairly translated and ready for evaluation, and $\beta_{ILM}$ and $\beta_{ELM}$ were robustly estimated.

However, in a real use case, we need a unique combination of $\beta_{ELM}$ and $\beta_{ILM}$ for each gender. 
Therefore, in Table \ref{tab:mean-beta} we report
the mean values of $\beta_{ELM}$ and $\beta_{ILM}$ over the 10 folds for each language 
pair.
We can notice that the optimal values are closely aligned across the three language directions.
In general, for M\textsubscript{B-ILM+ELM} $\beta_{ELM}$ is always higher than $\beta_{ILM}$.
Moreover,
another
clear and consistent trend emerging in all language pairs is the
necessity for higher $\beta_{ELM}$ and $\beta_{ILM}$ values when the speaker is female. In this condition, a higher contribution of the ELM is required to counterbalance the inherent bias of the base ST model towards masculine forms.

\subsection{Impact of $\beta_{ILM}$ and $\beta_{ELM}$}
\label{app:heatmaps}

In addition to empirically estimating $\beta_{ILM}$ and $\beta_{ELM}$ through cross-validation, 
we also investigated the importance of optimizing the balance between 
the ILM and the ELM
for mitigating 
bias
without compromising translation quality.
To this end, 
for each language direction
we computed the performance variations by adjusting $\beta_{ILM}$ and $\beta_{ELM}$ in increments of 0.05.
Figure \ref{fig:heatmap} shows BLEU and gender accuracy (calculated globally for F and M) scores
for each $(\beta_{ILM}, \beta_{ELM})$ combination.
Each heatmap defines a space bounded by the base ST model (bottom left corner: $(\beta_{ILM}, \beta_{ELM})=(0.0,0.0)$) and by the ST model with the ILM totally replaced by the gender-specific ELMs
(top right corner: $(\beta_{ILM}, \beta_{ELM})=(1.0,1.0)$).

The trends are similar for all the three language directions.
As for gender accuracy,
ELM integration appears to be more critical than ILM removal.
Specifically, we observe that the accuracy improves as the value of $\beta_{ELM}$ increases. 
Looking at BLEU,
we observe a diagonal ellipse-shaped trend with higher scores around the bottom left corner. 
This indicates that, to preserve translation quality,
$\beta_{ILM}$ and $\beta_{ELM}$ should be similar and not too high.
Overall,
although the trends for translation quality and gender accuracy differ, the two objectives share high results in the middle area.

Most importantly, we can notice that the results are not significantly affected by small variations in the weights, with wide smooth areas with similar scores and no isolated peaks. This demonstrates the robustness of our solution with respect to a suboptimal estimation of $\beta_{ILM}$ and $\beta_{ELM}$.

\begin{table*}[!htb]
\small
\centering
\begin{tabular}{llll}
\hline
\textbf{Lang.}  &  \textbf{Gender}  &  & \textbf{Example} \\ 
\hline

\multirow{5}{*}{en-es} & \multirow{5}{*}{F} & SRC & \begin{tabular}[c]{@{}l@{}}I felt \textbf{alienated}, \textbf{intimidated} and \textbf{judged} by many.\end{tabular} \\
 &  &  REF  & \begin{tabular}[c]{@{}l@{}}Me sentí \textbf{alienada}, \textbf{intimidada} y \textbf{juzgada} por muchos.\end{tabular} \\
 &  & M\textsubscript{B} & \begin{tabular}[c]{@{}l@{}}Me sentí \textbf{alienad\textcolor{teal}{a}}, \textbf{\textcolor{red}{intimidante}} (EN. \textit{intimidating}) y \textbf{juzgad\textcolor{teal}{a}} por muchos.\end{tabular} \\
 &  & M\textsubscript{B-ILM+ELM} & \begin{tabular}[c]{@{}l@{}}Me sentí \textbf{alienad\textcolor{teal}{a}}, \textbf{intimidad\textcolor{teal}{a}} y \textbf{juzgad\textcolor{teal}{a}} por muchos.\end{tabular} \\
 &  &  M\textsubscript{B+ELM} & \begin{tabular}[c]{@{}l@{}}Me sentí \textbf{\textcolor{red}{aislad}\textcolor{teal}{a}} (EN. isolated), \textbf{intimidad\textcolor{teal}{a}} y \textbf{juzgad\textcolor{teal}{a}} por muchos.\end{tabular} \\

\hline
\multirow{5}{*}{en-fr} & \multirow{5}{*}{F} & SRC & \begin{tabular}[c]{@{}l@{}}
I was \textbf{tired} of faking normal.\end{tabular} \\
&  & REF  & \begin{tabular}[c]{@{}l@{}}J'étais \textbf{fatiguée} de simuler la normalité.\end{tabular} \\ 
&  & M\textsubscript{B} & \begin{tabular}[c]{@{}l@{}}J'étais \textbf{fatigu\textcolor{red}{é}} d'avoir l'air normal.
\end{tabular} \\
&  & M\textsubscript{B-ILM+ELM} & \begin{tabular}[c]{@{}l@{}}J'étais \textbf{fatigu\textcolor{teal}{ée}} d'avoir l'air normal.\end{tabular} \\
&  &  M\textsubscript{B+ELM} & \begin{tabular}[c]{@{}l@{}}J'étais \textbf{fatigu\textcolor{red}{é}} \textcolor{red}{d'avoir l'impression d'être normal} (EN.\textit{ of having the impression of} \\ \textit{being normal}).\end{tabular} \\ 

\hline
\multirow{8}{*}{en-it} & \multirow{8}{*}{M} & SRC & \begin{tabular}[c]{@{}l@{}}In 2007, I was \textbf{hired} as a \textbf{curator} at the Denver Museum of Nature and Science.\end{tabular} \\
 &  & REF  & \begin{tabular}[c]{@{}l@{}}Nel 2007, fui \textbf{assunto} come \textbf{curatore} al Denver Museum of Nature and Science.\end{tabular} \\ 
 &  & M\textsubscript{B} & \begin{tabular}[c]{@{}l@{}}Nel 2007 sono stata \textbf{assunt\textcolor{red}{a}} come \textbf{cura\textcolor{teal}{tore}} al Museo d'\textcolor{red}{Arte Moderna di} Science \\ (EN. \textit{Modern Art of}).\end{tabular} \\
 &  & M\textsubscript{B-ILM+ELM} & \begin{tabular}[c]{@{}l@{}}Nel 2007 sono stato \textbf{assunt\textcolor{teal}{o}} come \textbf{cura\textcolor{teal}{tore}} al Museo d'\textcolor{red}{Arte Moderna di} Science \\ (EN. \textit{Modern Art of}).\end{tabular} \\
 &  &  M\textsubscript{B+ELM} & \begin{tabular}[c]{@{}l@{}}Nel 2007 sono stato \textbf{assunt\textcolor{teal}{o}} come \textbf{cura\textcolor{teal}{tore}} al Museo d'\textcolor{red}{Arte Moderna di Scienza} \\ (EN. \textit{of Modern Art of Science}).
\end{tabular} \\ 
\hline
\end{tabular}
\caption{Examples of outputs from the baseline M\textsubscript{B},  M\textsubscript{B-ILM+ELM} and M\textsubscript{B+ELM}, along with the corresponding source (SRC) and reference (REF).
We indicate the \textcolor{teal}{correct}/\textcolor{red}{wrong} gender translation for \textbf{words} on which gender accuracy is evaluated, as well as generic \textcolor{red}{mistranslations} of other \textit{words}.}
\label{tab:examples}
\end{table*}


\section{ST Model and Language Models}
\label{app:models}

\paragraph{ST Models}
Our direct ST models are made of a 12-layer Conformer~\cite{gulati20_interspeech} encoder, in light of its favorable results in ST \cite{inaguma-etal-2021-source},
and a 6-layer Transformer~\cite{transformer} decoder.
The architecture is also preceded by two 1D convolutional layers with 5 as kernel size and stride 2, as per \cite{wang2020fairseqs2t}.
We use 512 embedding features, 2,048 hidden features in the FFN, and a kernel size of 31 for Conformer convolutions.
In total, the ST models have 116M parameters.
We trained them with an auxiliary CTC loss on the 8th encoder layer \cite{gaido-etal-2022-efficient-yet}
and we leveraged the CTC module to compress the sequence length \cite{liu2020bridging,gaido-etal-2021-ctc}.
We encoded text into BPE \cite{sennrich-etal-2016-neural} using SentencePiece \cite{kudo-richardson-2018-sentencepiece} with a vocabulary size of 
8,000 \cite{di-gangi-etal-2020-target},
and we used Adam optimizer~\cite{DBLP:journals/corr/KingmaB14} ($\beta_1=0.9$, $\beta_2=0.98$) and Noam learning rate (lr) scheduler~\cite{transformer} (inverse square-root) starting from $0$ and reaching the 0.002 peak in $25,000$ warm-up steps.
The ST models for each language direction were trained for 50k steps on 4 NVIDIA A100 GPUs (40GB of RAM) with 40k tokens per mini-batch and 2 as update frequency, and we averaged the last 7 checkpoints.
To implement the specialized models (M\textsubscript{SP}), we fine-tuned
M\textsubscript{B} on the masculine/feminine partitions of the MuST-C data,
with a constant lr of 0.001 for 7 epochs, and we averaged the last 4 checkpoints.
All our models are implemented on fairseq \cite{ott-etal-2019-fairseq}.

\paragraph{Language Models}
The gender-specific ELMs are
Transformer decoders with 6 layers
(23M weights) trained with the same vocabularies and hyper-parameters of M\textsubscript{B}, except for the 
learning rate
warm-up updates that we set to 200.
We early stopped the training after 5 epochs without improvements on the validation loss, and we average the 5 checkpoints around the best on the validation set.

\begin{table*}[tb]
\centering
\small
\setlength{\tabcolsep}{5.5pt}
\begin{tabular}{l||cc|cc||cc|cc||cc|cc} 
\hline
\multirow{3}{*}{\textbf{Models}} & \multicolumn{4}{c||}{\textbf{en-es}} & \multicolumn{4}{c||}{\textbf{en-fr}} & \multicolumn{4}{c}{\textbf{en-it}} \\ 
\cline{2-13} 
& \multicolumn{2}{c|}{\textbf{Coverage}} & \multicolumn{2}{c||}{\textbf{Gender Acc.}} 
 & \multicolumn{2}{c|}{\textbf{Coverage}} & \multicolumn{2}{c||}{\textbf{Gender Acc.}} & \multicolumn{2}{c|}{\textbf{Coverage}} & \multicolumn{2}{c}{\textbf{Gender Acc.}} \\ 
 & M & F & M & F 
 & M & F & M & F 
 & M & F & M & F \\ 
\hline
M\textsubscript{B} & 
72.91 & \textbf{68.81} & 
\textbf{82.54} & 63.99 & 
\textbf{66.60} & 59.57 & 
\textbf{84.74} & 68.61 &
\textbf{58.55} & \textbf{60.30} & 
81.27 & 64.63 \\
M\textsubscript{SP} & 
\textbf{73.75} & 66.79 & 
82.48 & 65.83 & 
64.45 & 58.71 & 
83.51 & 69.18 & 
57.10 & 59.01 & 
\textbf{82.38} & \textbf{68.20} \\
M\textsubscript{B-ILM+ELM} & 
72.41 & \textbf{68.81} & 
81.40 & \textbf{66.59} & 
63.09\textsuperscript{a} & \textbf{59.78} & 
82.87 & \textbf{69.87} & 
57.74 & 56.87\textsuperscript{a} &
81.44 & 67.36\textsuperscript{A} \\
\hline
\end{tabular}
\caption{(Term) coverage ($\uparrow$) and 
M/F
gender accuracy (Gender Acc., $\uparrow$) scores for Category 2 of MuST-SHE.
\textsuperscript{A/a} and \textsuperscript{B/b} indicate that the improvement (uppercase) or the degradation (lowercase) of our technique over the baseline (M\textsubscript{B}) and the fine-tuning approach (M\textsubscript{SP}), respectively, 
is statistically significant (bootstrap resampling with 95\% CI, \citealt{koehn-2004-statistical}).}
\label{tab:cat2}
\end{table*}

\section{Examples}
\label{app:examples}
In Table \ref{tab:examples} we report output samples that well exemplify the behavior of our models and the baseline.

First, the examples in en-fr and en-it confirm the gender-accuracy improvements of our methods 
discussed
in \S\ref{sec:main_results}. The outputs of the baseline (M\textsubscript{B}) contain speaker-dependent words with the wrong gender, as a masculine form (fr: \textit{fatigué}, en: \textit{tired}) is used with a female speaker in en-fr, and a  feminine form (it: \textit{assunta}, en: \textit{hired}) with a male speaker in en-it. 
Our solution (M\textsubscript{B-ILM+ELM}), instead, consistently 
generates the correct gender inflection
in both cases
(fr: \textit{fatiguée} 
and 
it: \textit{assunto}), even without the ILM removal (M\textsubscript{B+ELM}). This is in line with the analysis in 
Appendix \ref{app:beta},
where we have seen
that gender accuracy mostly depends on ELM integration.

Looking at the en-es example, instead, M\textsubscript{B} correctly assigns the gender
but it wrongly translates
one of the adjectives referred to the speaker, using the epicene term \textit{intimidante} (en: \textit{intimidating}) for \textit{intimidated}. Similarly, the output of M\textsubscript{B+ELM}, 
although with the correct gender,
contains 
an error 
(\textit{alienated} is rendered as \textit{aislada}, en: \textit{isolated}). 
Instead, all adjectives are correct in the output of M\textsubscript{B-ILM+ELM}, confirming its higher coverage (see \S\ref{sec:main_results}) and the importance of ILM removal to avoid quality drops (see 
Appendix \ref{app:beta}
and the BLEU scores in \S\ref{sec:main_results}). The latter aspect also emerges from the errors introduced by M\textsubscript{B+ELM} with respect to M\textsubscript{B} both in en-fr and in en-it, which are not present in the output of M\textsubscript{B-ILM+ELM}: for instance, in en-fr, the translation of \textit{faking normal} alters it meaning, deviating to \textit{avoir l'impression d'être normal} (en: \textit{having the impression of being normal}).

\section{Impact on 
Human Referents Other than the Speaker}
\label{app:cat2}

Our work is dedicated to the gender translation of speaker-dependent words i.e., those words that refer to the first-person-singular referent. However, the improvements in handling this aspect should not come to the detriment of the accuracy in assigning the gender to referents different from the speaker. To ensure that this is not the case, we 
also evaluated
the gender translation on the ``Category 2'' of the MuST-SHE benchmark.
This
contains approximately 500 sentences with the annotation of words related to 
third-person references, 
whose gender is independent from that of the speaker. 
The results are presented in Table \ref{tab:cat2}.

As for
gender accuracy, we
observe that all systems are close for masculine forms (M), 
with 
variations that are not statistically significant.
The largest difference amounts to 1.87 points on en-fr between the baseline (M\textsubscript{B}) and our solution (M\textsubscript{B-ILM+ELM}). 
Similarly, M\textsubscript{B-ILM+ELM} and the specialized systems (M\textsubscript{SP}) achieve comparable scores on feminine forms (F)
while M\textsubscript{B} is constantly worse, with a statistically significant difference in en-it.

Looking at the term coverage, we do not see clear trends across language pairs. 
For 
F, 
M\textsubscript{B-ILM+ELM} suffers from a significant drop in en-it with respect to M\textsubscript{B}
while it achieves the best scores in en-es and en-fr. 
For 
M, 
there is a significant drop in en-fr, which is not confirmed in the other two language pairs.
In addition, the differences with M\textsubscript{SP} are always ascribable to random fluctuations.

All in all, we can conclude that
our debiasing solution specifically designed for speaker-dependent words
does not significantly alter
the gender assignment for referents different from the speaker.

\end{document}